\documentclass{article}
\usepackage[margin=0.72in]{geometry}
\usepackage{amsmath,amssymb,bm,graphicx,booktabs,tabularx,array,microtype}
\usepackage{caption,float,placeins,enumitem,needspace}
\usepackage{algorithm}
\usepackage{algpseudocode}

\usepackage[round,authoryear]{natbib}
\usepackage{xr-hyper}
\makeatletter
\newcommand*{\addFileDependency}[1]{\typeout{(#1)}\@addtofilelist{#1}\IfFileExists{#1}{}{\typeout{No file #1.}}}
\makeatother
\newcommand*{\myexternaldocument}[1]{\externaldocument[][nocite]{#1}\addFileDependency{#1.tex}\addFileDependency{#1.aux}}
\usepackage[hidelinks]{hyperref}
\myexternaldocument{appendix}
\graphicspath{{figures/}}
\setlist{itemsep=0.15em,topsep=0.25em}
\newcommand{\E}{\mathbb E}
\newcommand{\Var}{\operatorname{Var}}

\newcommand{\Tgrid}{\mathcal T}

\begin{document}
\begin{center}

{\LARGE\bfseries
{\Large\bfseries
Joint Modeling of Conditional Mean and Variability in\\[0.15cm]
Multilevel Functional Data: Application to\\[0.15cm]
Postprandial Glucose Trajectories in the AEGIS Study
\par}
\par}

\vspace{0.65cm}

{\large
Yang Xiao$^{1}$,
Francisco Gude$^{2,3}$,
Marcos Matabuena$^{1,*}$
\par}

\vspace{0.45cm}

{\small
$^{1}$ Division of Biological and Life Sciences,\\
Mohamed bin Zayed University of Artificial Intelligence,\\
Abu Dhabi, United Arab Emirates
\par}

\vspace{0.25cm}

{\small
$^{2}$ Instituto de Investigación Sanitaria de Santiago de Compostela (IDIS),\\
Santiago de Compostela, Spain
\par}

\vspace{0.25cm}

{\small
$^{3}$ Department of Medicine, University of Santiago de Compostela,\\
Santiago de Compostela, Spain
\par}

\vspace{0.4cm}

{\small
$^{*}$\textit{Corresponding author:}
\href{mailto:marcos.matabuena@mbzuai.ac.ae}
{marcos.matabuena@mbzuai.ac.ae}
\par}

\end{center}

\vspace{0.5cm}

\begin{abstract}
Functional data analysis (FDA) regression models provide a natural analytical framework for quantifying time-varying associations between Euclidean predictors and high-frequency biological signals (e.g., minute-by-minute physical activity measurements and glucose trajectories). In digital health applications, multiple functional trajectories are often collected from each participant. Existing multilevel functional regression methods primarily focus on modeling the conditional mean as a function of individual-level covariates (e.g., age, body mass index, and demographic characteristics), whereas conditional variability has received comparatively little attention, despite its importance for understanding biological processes and characterizing individual heterogeneity. To address this gap, we introduce, to the best of our knowledge, the first heteroskedastic functional mixed-effects regression framework that jointly models the conditional mean and variance of repeated functional observations while incorporating participant-specific random effects in both components. This framework enables the simultaneous characterization of time-varying biological responses and their variability, capturing dimensions of individual heterogeneity that conventional multilevel functional regression models cannot adequately represent. We further develop a participant-level bootstrap procedure to quantify uncertainty in the estimated functional parameters. We apply the proposed framework to postprandial glucose trajectories from the AEGIS study, revealing clinically relevant patterns of time-varying variability that are overlooked by conventional analyses focusing primarily on mean glucose responses. In this general adult population, premeal glucose levels, carbohydrate intake, and meal type exhibit distinct time-dependent associations with postprandial glucose variability, whereas glycated hemoglobin (HbA1c) shows a comparatively weaker association with variability despite retaining its association with the conditional mean. We also identify substantial between-participant heterogeneity in residual variability, even among individuals with similar observed covariates, highlighting dimensions of biological heterogeneity that cannot be adequately characterized through conditional mean models alone. Overall, our findings establish heteroskedastic functional mixed-effects regression as a flexible framework for jointly characterizing the conditional mean, variability, and participant-specific heterogeneity of longitudinal biological signals, with important implications for personalized modeling in digital health and precision nutrition in the analyzed case study.

\end{abstract}
\section{Introduction}

\begin{figure}[h]
\centering
\includegraphics[width=\linewidth,height=0.67\textheight,keepaspectratio]{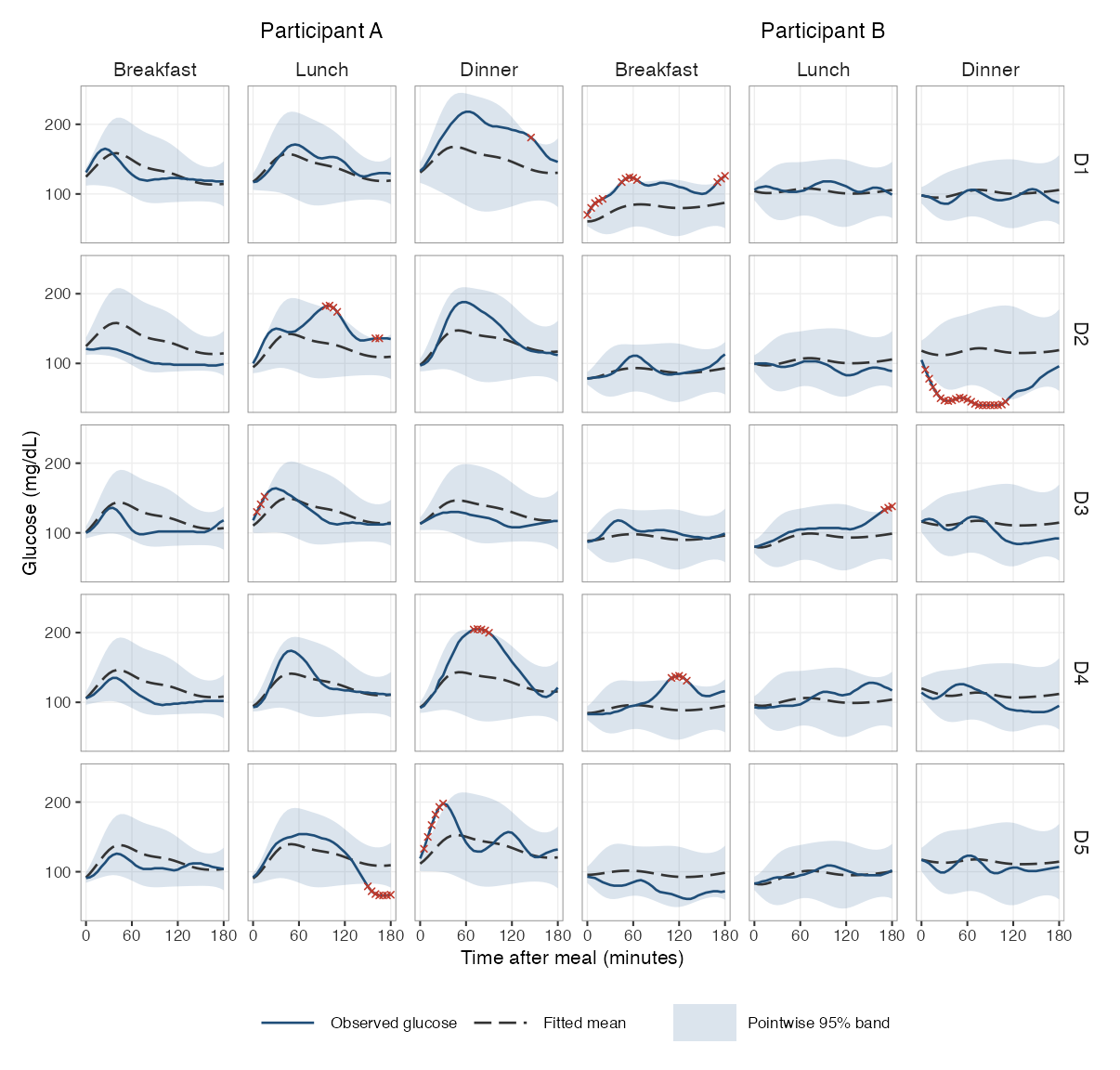}
\caption{Postprandial glucose trajectories of two AEGIS participants after breakfast, lunch and dinner on five study days (D1--D5), on a five-minute grid from the start of each meal. Solid lines are the observed glucose and dashed lines the participant's fitted mean for that meal. Shading is an approximate pointwise 95\% prediction band with asymptotic approximation, the fitted mean $\pm1.96$ times the fitted residual standard deviation; crosses mark observations outside it.}
\label{fig:fig_intro_data_structure}
\end{figure}

The ongoing revolution in biological measurement technologies is enabling the collection of multiple high-frequency time series of physiological variables, such as glucose levels and heart rate, on different time scales, with the potential to accelerate scientific discovery and improve clinical decision making \cite{Matabuena03042023}. A key feature of these modern datasets is that devices record observations every few seconds or minutes; these time-stamped measurements can be naturally represented as a mathematical function over time. Because individuals are typically monitored repeatedly (e.g., across days or other time periods), the resulting data structure collected is functional and multilevel (hierarchical) \cite{matabuena2026multilevel}. To emphasize this structure, Figure~\ref{fig:fig_intro_data_structure} displays, for two participants in the AEGIS study, glucose trajectories after breakfast, lunch and dinner, recorded every five minutes on five days, together with the corresponding fitted mean curves and gaussian assympotic prediction bands. The plots of individual glucose trajectories show substantial heterogeneity both within and between individuals across meals and over days, underscoring the importance of multilevel random-effects models that capture individual variation specific to the time after a meal for personalized prediction.

In digital health applications, multilevel functional models are increasingly used to analyze high-frequency measurements of functional biomarkers. When individual-level covariate information is not available and only repeated functional trajectories are observed, the prevailing approach is multilevel functional principal component analysis (MFPCA)~\citep{di2009multilevel}. MFPCA has been developed for a wide range of hierarchical designs and estimation paradigms~\citep{crainiceanu2024book}, and recent work has introduced computationally efficient algorithms for both Gaussian and non-Gaussian functional outcomes~\citep{cui2023fast,leroux_fast_2023}.

When individual-level covariate information is available (e.g. demographic and anthropometric measures), a range of frequentist and Bayesian functional regression methods have been proposed to model the conditional mean function~\citep{morris2006wavelet,greven2017general}. Scalable implementations for large cohorts, such as NHANES, have also emerged~\citep{cui2022fast,sun2025ultra}. These often rely on a ``mass-univariate-then-smooth'' pipeline: separate univariate models are fitted at each time point on the functional grid and the resulting coefficient estimates are then smoothed over time to enforce temporal regularity and improve statistical inference. Because each time point can be modeled independently, this strategy is naturally parallelizable and scales well in large epidemiological studies---for example, minute-by-minute accelerometer-derived physical activity signals collected from about 100,000 participants in the UK Biobank.

Despite substantial progress in multilevel functional modeling, most multilevel functional regression methods target only the conditional mean, leaving other distributional features, such as the conditional variance, largely unmodeled. Approaches that capture temporal structure in conditional variability are still scarce \citep{Backenroth03072018,stoecker2021boosting}, although variability is key for characterizing individual heterogeneity and for designing temporally targeted digital health interventions (e.g., chronobiologically personalized interventions). For scalar repeated measurements, mixed-effects location–scale models allow covariates and subject-specific random effects to influence both the conditional mean and within-subject variance~\citep{hedeker2008application,hedeker2013mixregls}. Subsequent developments have extended these models to nonlinear learning curves~\citep{williams2019nonlinear} and dyadic data~\citep{rast2018dyadic}, while recent computational advances have enabled more efficient estimation~\citep{gill2023fast,wu2026variational}. However, these approaches are designed for scalar responses and do not directly accommodate the functional structure of repeated trajectories, where both the conditional mean and variability may evolve continuously over time. To address this limitation, we propose a joint multilevel functional location–scale model that simultaneously estimates covariate-dependent mean and variance functions, enabling the characterization of time-varying biological responses and their heterogeneity across individuals.

\paragraph{Why variability matters.}

Understanding biological variability is essential for elucidating physiological mechanisms, characterizing individual heterogeneity, and informing the development of personalized, time-dependent interventions. Across clinical domains, variability provides information beyond average responses. In oncology, variability patterns have been used to characterize tumor heterogeneity and identify clinically relevant subtypes~\citep{doi:10.1073/pnas.1304376110}; in diabetes, glucose variability complements mean glucose as an indicator of glycemic regulation and metabolic instability~\citep{rodbard2018glucose}; and in physical activity, temporal variability has been associated with circadian organization and age~\citep{xiao2015quantifying,li2021demographic}. Despite its clinical relevance, biological variability is frequently summarized using scalar measures, such as daily standard deviations~\citep{rodbard2018glucose}, which obscure how variability evolves throughout the day or in response to specific physiological stimuli. This limitation is particularly relevant for high-frequency biological signals, such as continuous glucose monitoring (CGM) trajectories, where both the magnitude and timing of variability may carry important physiological information. For example, two individuals may exhibit similar average postprandial glucose responses but differ substantially in the day-to-day variability of their glucose trajectories, with these differences changing over the postprandial period (Figure~\ref{fig:fig_intro_data_structure}). Characterizing such heterogeneity requires moving beyond conditional mean regression to explicitly model the conditional variance as a function of time and individual-level covariates, thereby capturing aspects of biological regulation that conventional summary measures may overlook.

\paragraph{AEGIS motivation.}
Our methodological development is motivated by the need to characterize time-dependent variability in postprandial glucose responses among adults without diabetes, using continuous glucose monitoring (CGM) data from the AEGIS study~\citep{gude2017glycemic,calvomalvar2025age}, while accounting for individual characteristics and meal composition. Postprandial glucose regulation reflects complex interactions among dietary composition, insulin sensitivity, pancreatic beta-cell function, and other physiological processes. Large-scale studies have documented substantial interindividual heterogeneity in glycemic responses to identical meals, providing a scientific basis for personalized nutrition approaches~\citep{zeevi2015personalized,berry2020human}. More recently, detailed metabolic phenotyping has demonstrated that individual responses to different carbohydrate-rich foods are associated with underlying metabolic characteristics, including insulin resistance and beta-cell dysfunction~\citep{wu2025individual}. However, considerable variability may also arise within the same individual across repeated exposures to similar meals. Recent evidence has highlighted the limited reproducibility of postprandial glycemic responses to duplicate meals, raising questions about the reliability of personalized dietary recommendations derived from a small number of CGM observations~\citep{hengist2025imprecision}. Together, these findings motivate statistical approaches that distinguish expected postprandial glucose responses from their within-individual variability, while accounting for individual metabolic characteristics and differences in meal composition.

Although glycemic variability is increasingly recognized as an important dimension of glucose regulation, clinical assessments commonly rely on scalar summaries, such as the coefficient of variation and time-in-range metrics~\citep{battelino2019clinical}. While these measures provide useful information about overall glycemic patterns, they do not directly characterize how the conditional variability of repeated postprandial trajectories evolves over the period following a meal. Functional data analysis (FDA) offers a natural framework for studying high-resolution glucose trajectories~\citep{gaynanova2020modeling}, and previous analyses of AEGIS postprandial responses have employed multilevel functional regression to investigate associations with individual and meal characteristics~\citep{calvomalvar2025age,matabuena2026beyond}. However, existing approaches have primarily focused on conditional mean responses or summarized variability using scalar measures~\citep{gude2017glycemic}. Consequently, the time-dependent structure of within-individual glucose variability, and its association with metabolic and dietary characteristics, remains insufficiently characterized.

\paragraph{Goals and contributions.}
This paper addresses the lack of methods for estimating conditional variability in multilevel functional data. We propose a computationally efficient algorithm that estimates the conditional mean and then, in further steps, the conditional variability (heteroscedasticity) in multilevel functional models and provides  uncertainty quantification for functional parameters. The model is a functional mixed-effects location--scale model, in which covariates and participant random functions act on both the mean and the log residual standard deviation of each trajectory (Section~\ref{sec:model}). The conditional mean and the conditional variability at a covariate profile are two parallel primary targets, each with its own time-varying covariate effects (Section~\ref{sec:targets}). Our key idea is a three-step fitting strategy: (i) to estimate the conditional mean with a multilevel functional regression fitted score by score on a spline basis, (ii) to estimate the conditional variability by a regression of cross-fitted squared residuals on covariates, with participant random functions, and (iii) to correct this scale fit for the error in predicting each participant's mean. Step (iii) addresses the main difficulty of variance estimation in multilevel functional data: when participants contribute few trajectories, the squared residuals also contain the error of the predicted participant mean, which pulls the estimated variance slopes towards zero and does not vanish as the number of participants grows (Section~\ref{sec:corrected}). The correction estimates this component, removes it through conditional refits of the scale model, and estimates the level of variability separately. To propagate uncertainty from all steps, we use a participant bootstrap that repeats every step and yields  pointwise and simultaneous confidence bands  for the conditional mean and conditional variance functional parameters.

We further demonstrate the practical relevance of our framework through an analysis of postprandial glucose trajectories from the AEGIS study, uncovering distinct time-varying associations between premeal glucose levels, carbohydrate intake, meal type, and conditional glucose variability. These associations would be systematically underestimated without the proposed bias correction. Finally, we introduce an individual-level functional $R^{2}$ that quantifies the model-based reduction in squared prediction error relative to a common population variance. Its substantial variation across participants reveals pronounced heterogeneity in the extent to which the model explains individual glucose variability.

\paragraph{Paper roadmap.}

Section~\ref{sec:methods} introduces the modeling framework and the associated estimation and inference procedures.
Section~\ref{sec:application} presents a detailed analysis of the AEGIS case study.
Section~\ref{sec:discussion} concludes with a discussion of the results, limitations, and future work. The Appendix gives estimation and inference details and implementation details of the case study.

%biology, captures continuos biological ppatenr , multiple measures per individuos

%temporal dimensional of biologicalmenete discover, and allow to personalizze the interventions

%examples in biological signals 

%show the mean for two individuals and the variations of the variability . Two individuals simialar mean different variability, differences in the time

%our papers 

%Importantc eof variance and cronobiology and conbient totehre

%examplify for the case study

%multilevel funcitonal models regreession and unsupervised and cluter

%Contributions

%emphaziee
%use time series to coonnect indivudyals models foer globals

%\newpage

\section{Methodology}
\label{sec:method}
\label{sec:methods}

We model the conditional mean and variability of repeated functional trajectories with a multilevel location--scale model. Estimation fits the mean, estimates the scale from cross-fitted residuals and corrects it for error in the predicted participant mean effects; inference resamples participants and repeats every estimation step.

\subsection{Data, notation and model}
\label{sec:model}

We observe $N$ independent participants sampled from a common population. Participant $i\in [N]:=\{1,\dots, N\}$ has available $n_i\geq 2$ functional observations $Y_{ij}(\cdot)\in\mathcal{H}$, $j=1,\ldots,n_i$, where $\mathcal{H}$ is a separable Hilbert space of real-valued functions defined on a common time domain $\mathbb{T}$. A natural example that we assume along this paper is $\mathcal{H}=L^2(\mathbb{T})$, the space of square-integrable functions, equipped with the inner product $\langle f,g\rangle_{\mathcal{H}}=\int_{\mathbb{T}}f(t)g(t)\,dt$ and the induced norm $\|f\|_{\mathcal{H}}=\{\int_{\mathbb{T}}f^2(t)\,dt\}^{1/2}$. In our application, $\mathbb{T}=[0,180]$ represents the first 180 minutes following a meal. The trajectories are observed on a common discrete grid $\Tgrid=\{t_1,\ldots,t_T\}\subset\mathbb{T}$.

Each functional observation is associated with a $p$-dimensional covariate vector $X_{ij}=(1,X_{ij,-}^{\top})^{\top}\in\mathbb{R}^p$, where the first component corresponds to the intercept. We write $X_i=(X_{i1},\ldots,X_{in_i})$ for the collection of covariates associated with participant $i$. Covariates may vary across participants, across repeated trajectories within the same participant, or both.

 Grid points and basis functions are indexed by $k=1,\ldots,T$
and $\ell$, respectively, and $M=\sum_{i=1}^{N}n_i$ denotes
the total number of observed trajectories. We use
$Y_{ij}(\cdot)$ to denote the entire functional trajectory
and $Y_{ij}(t)$ to denote its value at time $t$.
For covariate vectors and coefficient functions, the subscript
$-$ indicates exclusion of the intercept component, whereas
the subscript $a$ identifies the $a$th component.
The superscripts $\mu$ and $\omega$ distinguish quantities
associated with the mean and scale models, respectively,
including the spline bases $B_\ell^{(\mu)}(\cdot)$ and
$B_\ell^{(\omega)}(\cdot)$ and their respective dimensions
$K_\mu$ and $K_\omega$.
Hats denote estimators, while the superscript $(-f)$
indicates quantities computed excluding cross-fitting fold $f$,
$(r)$ denotes the $r$th scale-model refit ($r=0$ for the
initial fit), and $*$ indicates bootstrap quantities obtained
by resampling participants.
Finally, $I_n$ denotes the $n\times n$ identity matrix,
$\mathbf{1}_n$ the $n$-dimensional vector of ones, and
$\perp\!\!\!\perp$ statistical independence.

Suppose that the trajectories follow a function-on-scalar
location--scale model:
\begin{equation}
Y_{ij}(t)
=
X_{ij}^{\top}\beta(t)+u_i(t)
+
\exp\{X_{ij}^{\top}\gamma(t)+v_i(t)\}
\epsilon_{ij}(t),
\qquad
i=1,\ldots,N,\quad
j=1,\ldots,n_i,\quad
t\in\Tgrid,
\label{eq:model}
\end{equation}
where $\beta(\cdot)$ and $\gamma(\cdot)$ are coefficient
functions governing the conditional mean and log residual
standard deviation (SD), respectively.
The participant-specific random effects $u_i(\cdot)$
and $v_i(\cdot)$ act on the mean and log residual SD
and are shared across all trajectories of participant $i$.
Both effects are centered, with
$\E\{u_i(t)^2\}<\infty$ and
$\E\{\exp(2v_i(t))\}<\infty$.
When the functional quantities are replaced by scalars,
model~\eqref{eq:model} reduces to a mixed-effects
location--scale model for repeated measurements
\citep{hedeker2008application,hedeker2013mixregls,lee2006double}.
In the absence of participant-specific random effects,
it reduces to a location--scale regression model
\citep{rigby2005gamlss}.

Define
\[
\mu_{ij}(t)
=
X_{ij}^{\top}\beta(t)+u_i(t),
\qquad
\omega_{ij}(t)
=
\exp\{X_{ij}^{\top}\gamma(t)+v_i(t)\},
\]
as the participant-specific conditional mean and residual
SD, respectively. Thus,
$Y_{ij}(t)=\mu_{ij}(t)+\omega_{ij}(t)\epsilon_{ij}(t)$,
where $\epsilon_{ij}(t)$ is a standardized error process.
We impose the following assumptions:
\begin{enumerate}[label=(C\arabic*),leftmargin=*]
\item
$\E\{\epsilon_{ij}(t)\mid X_i,u_i,v_i\}=0$
and
$\Var\{\epsilon_{ij}(t)\mid X_i,u_i,v_i\}=1$;

\item
$\epsilon_{i1}(\cdot),\ldots,\epsilon_{in_i}(\cdot)$
are conditionally independent given $(X_i,u_i,v_i)$;

\item
$\{u_i(\cdot),v_i(\cdot)\}$ is independent of
$(n_i,X_i)$.
\end{enumerate}

The random effects capture participant-specific departures
from the population-level mean and scale functions.
For participants with the same covariates, a positive
$u_i(t)$ indicates a higher expected glucose level at
time $t$ after a meal, whereas a positive $v_i(t)$
indicates greater residual variability around the
participant-specific mean trajectory relative to the
fixed-effect scale.
Conditions (C1) and (C2) specify that, conditional on
the covariates and participant effects, trajectories
are independent, with conditional means
$\mu_{ij}(\cdot)$ and residual SDs $\omega_{ij}(\cdot)$.
Thus, within-participant dependence across trajectories
is induced by the shared random effects.

Condition (C3) assumes that the participant effects
are independent of both the covariates and the number
of observed trajectories, $n_i$
\citep{seaman2014review}.
This assumption allows the coefficient functions
$\beta(\cdot)$ and $\gamma(\cdot)$ to characterize
covariate associations with the conditional mean
and log residual SD, as developed in
Section~\ref{sec:targets}.
Independence from $n_i$ is particularly relevant because
participants contributing more trajectories receive
greater weight in estimation; an association between
cluster size and the random effects could therefore
introduce bias.
The mean and scale random effects may be correlated,
as in scalar mixed-effects location--scale models
\citep{hedeker2008application}.
Importantly, these assumptions impose no restrictions
on temporal dependence within individual trajectories.

\subsection{Estimands}
\label{sec:targets}

Our primary estimands characterize the conditional mean and
within-participant variability of functional trajectories
at a given covariate profile $x$. Specifically, we define
the population-level mean function
$m(x,t)=x^{\top}\beta(t)$ and the participant-averaged
residual variance
\begin{equation}
\begin{aligned}
\bar q(x,t)
&=\E\left[
\Var\{Y_{ij}(t)\mid X_{ij},u_i,v_i\}
\mid X_{ij}=x
\right]\\
&=\E\{\omega_{ij}^2(t)\mid X_{ij}=x\}\\
&=\exp\{2x^{\top}\gamma(t)\}
  \E\left[\exp\{2v_i(t)\}\right].
\end{aligned}
\label{eq:reference}
\end{equation}
The second equality follows from (C1), whereas the third
uses (C3). The mean function $m(x,t)$ represents the
expected trajectory at covariate profile $x$, averaged
over participant-specific effects. In contrast,
$\bar q(x,t)$ quantifies the residual variability
around participant-specific mean trajectories, averaged
over participants with the same covariate profile.
We call $m(x,\cdot)$ the reference mean, and
$m(x,\cdot)$ and $\bar q(x,\cdot)$ together the
reference moments at profile $x$.

Under (C1) and (C3), the total conditional variance
admits the decomposition
\[
\Var\{Y_{ij}(t)\mid X_{ij}=x\}
=
\Var\{u_i(t)\}+\bar q(x,t),
\]
separating between-participant heterogeneity from
within-participant residual variability.

The non-intercept coefficient functions describe
covariate associations with the mean and residual
variability. Holding other covariates fixed, a one-unit
increase in the $a$th covariate changes the conditional
mean by $\beta_a(t)$ and multiplies the
participant-specific residual SD by
$\exp\{\gamma_a(t)\}$.
These effects describe conditional associations rather
than causal relationships.

Let $\gamma_-(\cdot)$ denote the non-intercept
coefficient functions in the scale model.
Since
\[
\frac{1}{2}\log\bar q(x,t)
=
x^{\top}\gamma(t)
+
\frac{1}{2}\log
\E\left[\exp\{2v_i(t)\}\right],
\]
averaging over the participant-specific scale effects
changes only the intercept, leaving the covariate slopes
$\gamma_-(\cdot)$ unchanged.
Accordingly, we report the overall level of
within-participant variability through
$\bar q(x,\cdot)$ rather than through the intercept
of the log-SD model.
Appendix~\ref{app:targets} further relates
$\bar q(x,\cdot)$ to the residual variance
evaluated at $v_i(\cdot)=0$.

\subsection{Estimating the mean and the scale}
\label{sec:estimation}
\subsubsection{Conditional mean}
\label{sec:mean}
We represent the coefficient functions in a finite spline basis and estimate participant effects through score-wise linear mixed models, following functional mixed-model constructions~\citep{guo2002functional,greven2017general} and basis-space functional mixed models that fit each basis coefficient separately~\citep{morris2006wavelet}. 
Let $B_1^{(\mu)}(\cdot),\ldots,B_{K_\mu}^{(\mu)}(\cdot)$ be a cubic B-spline basis for the mean model, made orthonormal on $\Tgrid$ by QR decomposition: with $\bm B^{(\mu)}(t)=\{B_1^{(\mu)}(t),\ldots,B_{K_\mu}^{(\mu)}(t)\}^{\top}$, $T^{-1}\sum_{k=1}^T\bm B^{(\mu)}(t_k)\bm B^{(\mu)}(t_k)^{\top}=I_{K_\mu}$. We approximate
\[
\beta(t)\approx\sum_{\ell=1}^{K_\mu}\theta_\ell B_\ell^{(\mu)}(t),\qquad
u_i(t)\approx\sum_{\ell=1}^{K_\mu}a_{i\ell}B_\ell^{(\mu)}(t),
\]
with projection coefficients $\theta_\ell=T^{-1}\sum_{k=1}^T\beta(t_k)B_\ell^{(\mu)}(t_k)\in\mathbb R^p$ and $a_{i\ell}=T^{-1}\sum_{k=1}^Tu_i(t_k)B_\ell^{(\mu)}(t_k)$. Projecting~\eqref{eq:model} in the same way gives exactly
\begin{equation}
C_{ij\ell}=X_{ij}^{\top}\theta_\ell+a_{i\ell}+\xi_{ij\ell},
\qquad i=1,\ldots,N,\quad j=1,\ldots,n_i,\quad \ell=1,\ldots,K_\mu,
\label{eq:score-lmm}
\end{equation}
where $C_{ij\ell}=T^{-1}\sum_{k=1}^T Y_{ij}(t_k)B_\ell^{(\mu)}(t_k)$ is the projection score of trajectory $j$ in direction $\ell$ and $\xi_{ij\ell}=T^{-1}\sum_{k=1}^T\omega_{ij}(t_k)\epsilon_{ij}(t_k)B_\ell^{(\mu)}(t_k)$ is the projected residual; basis approximation error enters only through the reconstruction in~\eqref{eq:muhat-compact}. For each $\ell$, \eqref{eq:score-lmm} is a random-intercept model, with $a_{i\ell}$ shared by the trajectories of participant $i$ and $\xi_{ij\ell}$ of mean zero (Appendix~\ref{app:reml}). It is fitted by restricted maximum likelihood (REML)~\citep{bates2015lme4} as a working Gaussian mixed model with $a_{i\ell}\sim N(0,\tau_\ell^2)$ and $\xi_{ij\ell}\sim N(0,\sigma_\ell^2)$; cross-score covariance is not modeled in these fits. The fitted coefficients give
\begin{equation}
\widehat\beta(\cdot)=\sum_{\ell=1}^{K_\mu}\widehat\theta_\ell B_\ell^{(\mu)}(\cdot),\qquad
\widehat u_i(\cdot)=\sum_{\ell=1}^{K_\mu}\widehat a_{i\ell}B_\ell^{(\mu)}(\cdot),
\label{eq:muhat-compact}
\end{equation}
where $\widehat a_{i\ell}$ is the shrinkage prediction of participant $i$'s effect from that participant's trajectories~\citep{harville1976extension}, and $\widehat\mu_{ij}(t)=X_{ij}^{\top}\widehat\beta(t)+\widehat u_i(t)$. Appendix~\ref{app:reml} gives the basis construction, the derivation and properties of~\eqref{eq:score-lmm} and the REML fit.

\subsubsection{Conditional scale}
\label{sec:scale}
The residual $Y_{ij}(t)-\widehat\mu_{ij}(t)$ of Section~\ref{sec:mean} is on average smaller in magnitude than the deviation $Y_{ij}(t)-\mu_{ij}(t)$, because $Y_{ij}(\cdot)$ itself enters $\widehat u_i(\cdot)$ and pulls $\widehat\mu_{ij}(\cdot)$ towards $Y_{ij}(\cdot)$. With three trajectories, for instance, each makes up a third of the average from which $\widehat u_i(\cdot)$ is predicted.

We therefore use cross-fitted residuals~\citep{fan2012variance,chernozhukov2018double}, which replace $\widehat\mu_{ij}(\cdot)$ by a prediction from a mean fit that does not use $Y_{ij}(\cdot)$. The trajectories are split into folds $\mathcal I_1,\ldots,\mathcal I_F$, $F=\max_i n_i$, each containing at most one trajectory per participant. For $(i,j)\in\mathcal I_f$, refitting the mean model without fold $f$ gives $\widehat\mu_{ij}^{(-f)}(t)=X_{ij}^{\top}\widehat\beta^{(-f)}(t)+\widehat u_i^{(-f)}(t)$, where $\widehat u_i^{(-f)}(\cdot)$ is predicted from participant $i$'s other trajectories; the cross-fitted residual and its square are
\begin{equation}
\widehat e_{ij}^{(-f)}(t)=Y_{ij}(t)-\widehat\mu_{ij}^{(-f)}(t),
\qquad Q_{ij}(t)=\widehat e_{ij}^{(-f)}(t)^2.
\label{eq:cf-residual}
\end{equation}
Each trajectory lies in exactly one fold, so $f$ is determined by $(i,j)$; we omit the superscript $(-f)$ from $Q_{ij}(t)$ and from later quantities computed from the cross-fitted fits.

We model the conditional mean of $Q_{ij}(\cdot)$ with a log-link Gamma regression, a standard approach to variance-function estimation from squared residuals~\citep{davidian1987variance}:
\begin{equation}
\eta_{ij}(t)=\log\E\{Q_{ij}(t)\mid X_{ij},b_i\}
=X_{ij}^{\top}\alpha(t)+b_i(t),
\label{eq:scale-working}
\end{equation}
where $\alpha(\cdot)$ contains the covariate effects on this log mean and $b_i(\cdot)$ is a participant random function. The Gamma distribution is a working assumption for estimating this mean. As in Section~\ref{sec:mean}, both functions are expanded in an orthonormalized cubic B-spline basis $B_1^{(\omega)}(\cdot),\ldots,B_{K_\omega}^{(\omega)}(\cdot)$:
\[
\alpha(t)=\sum_{\ell=1}^{K_\omega}\zeta_\ell B_\ell^{(\omega)}(t),\qquad
b_i(t)=\sum_{\ell=1}^{K_\omega}b_{i\ell}B_\ell^{(\omega)}(t).
\]
The participant coefficients $b_{i\ell}$ are penalized as random effects, with one penalty for each $\ell$ selected by fast REML, treating the grid points of a trajectory as independent in this working fit (Appendix~\ref{app:gamma}). Like $\theta_\ell$, the coefficients $\zeta_\ell$ are not penalized, so the smoothness of $\beta(\cdot)$ and $\alpha(\cdot)$ is set by $K_\mu$ and $K_\omega$, which are fixed before fitting.

If the mean-prediction error $\mu_{ij}(t)-\widehat\mu_{ij}^{(-f)}(t)$ were zero, then $\E\{Q_{ij}(t)\mid X_{ij},v_i\}=\omega_{ij}(t)^2$ by (C1), so $\eta_{ij}(t)$ in~\eqref{eq:scale-working} would equal $\log\omega_{ij}(t)^2=2\{X_{ij}^{\top}\gamma(t)+v_i(t)\}$. Matching this with $X_{ij}^{\top}\alpha(t)+b_i(t)$ gives $\alpha(t)=2\gamma(t)$ and $b_i(t)=2v_i(t)$, and hence the estimates
\begin{equation}
\widehat\gamma^{(0)}(t)=\tfrac12\widehat\alpha^{(0)}(t),\qquad
\widehat v_i^{(0)}(t)=\tfrac12\widehat b_i^{(0)}(t).
\label{eq:omegahat}
\end{equation}
However, the mean-prediction error is not zero, so these estimates are uncorrected; the superscript $(0)$ marks them as the starting point of the correction in Section~\ref{sec:corrected}. This two-stage construction draws on separate dispersion modeling~\citep{smyth1989generalized} and functional and distributional regression~\citep{stoecker2021boosting,klein2024distributional}.

\subsection{Scale correction for mean-prediction error}
\label{sec:corrected}
By~\eqref{eq:model} and~\eqref{eq:cf-residual}, a cross-fitted residual is the error $\omega_{ij}(t)\epsilon_{ij}(t)$ of the held-out trajectory plus the mean-prediction error $\Delta_{ij}(t)=\mu_{ij}(t)-\widehat\mu_{ij}^{(-f)}(t)$:
\[
\widehat e_{ij}^{(-f)}(t)=\omega_{ij}(t)\epsilon_{ij}(t)+\Delta_{ij}(t),\qquad
\Delta_{ij}(t)=\{u_i(t)-\widehat u_i^{(-f)}(t)\}+X_{ij}^{\top}\{\beta(t)-\widehat\beta^{(-f)}(t)\}.
\]
As $N$ grows, the second term of $\Delta_{ij}(\cdot)$ vanishes at rate $N^{-1/2}$, apart from the spline approximation error of $\beta(\cdot)$, because $\widehat\beta^{(-f)}(\cdot)$ is estimated from all participants. The first term does not vanish when $n_i$ stays bounded, because $\widehat u_i^{(-f)}(\cdot)$ is predicted from only the $n_i-1$ other trajectories of participant $i$. 
Under (C1) and (C2), given $(X_i,u_i,v_i)$, $\epsilon_{ij}(\cdot)$ has mean zero and is independent of the training trajectories that determine $\Delta_{ij}(\cdot)$, so the cross term $2\omega_{ij}(t)\epsilon_{ij}(t)\Delta_{ij}(t)$ has mean zero. Squaring and taking expectations gives (Appendix~\ref{app:targets})
\begin{equation}
\E\{Q_{ij}(t)\mid X_i,v_i\}
=\omega_{ij}(t)^2+\psi_{ij}(t)+O(N^{-1}),\qquad
\psi_{ij}(t)=\E[\{u_i(t)-\widehat u_i^{(-f)}(t)\}^2\mid X_i,v_i].
\label{eq:pev-moment}
\end{equation}
Unlike the working model~\eqref{eq:scale-working}, which uses $X_{ij}$ only, this expectation conditions on all of $X_i$, because $\widehat u_i^{(-f)}(\cdot)$ depends on the covariates of participant $i$'s other trajectories; it averages over $u_i(\cdot)$ and the training errors. 
On average, the squared residual therefore exceeds $\omega_{ij}(t)^2$ by the prediction-error variance $\psi_{ij}(t)$. 

For a trajectory-level covariate, which changes $\omega_{ij}(t)^2$ but leaves $\psi_{ij}(t)$ unchanged, the local log-variance slope is multiplied by $\omega_{ij}(t)^2/\{\omega_{ij}(t)^2+\psi_{ij}(t)\}<1$, so the uncorrected slopes are pulled towards zero (Appendix~\ref{app:targets}). The distortion is largest for participants with few trajectories, whose effects are predicted least precisely. The correction estimates $\psi_{ij}(\cdot)$ (Step~1), refits the scale model with this component removed to estimate the slopes (Step~2), and then estimates the level of $\bar q(x,\cdot)$ (Step~3).

\emph{Step 1: estimate the prediction-error variance.}
Let $a_i=(a_{i1},\ldots,a_{iK_\mu})^{\top}$ and $\xi_{ij}=(\xi_{ij1},\ldots,\xi_{ijK_\mu})^{\top}$ collect the participant scores and projected residuals in~\eqref{eq:score-lmm}.
In the training set of fold $f$, participant $i$ has $n_i-1$ trajectories, and the shrinkage factors of Appendix~\ref{app:reml} give
\[
L_i=\operatorname{diag}_\ell\Bigl\{\frac{\widehat\tau_\ell^2}{\widehat\tau_\ell^2+\widehat\sigma_\ell^2/(n_i-1)}\Bigr\},\qquad
u_i(t)-\widehat u_i^{(-f)}(t)\approx\bm B^{(\mu)}(t)^{\top}\{(I-L_i)a_i-L_i\bar\xi_i\},
\]
where $\bar\xi_i$ is the average of the $\xi_{ij'}$ over these $n_i-1$ trajectories, and the approximation ignores the estimation error in $\widehat\theta_\ell^{(-f)}$ and the part of $u_i(\cdot)$ outside the mean basis (Appendix~\ref{app:targets}). The two terms are the error from shrinkage towards the population mean and the noise in the average of the training trajectories~\citep{harville1976extension}; by (C1) they are uncorrelated, so $\psi_{ij}(t)$ is approximately the sum of their variances. With $\widehat\Sigma_a$ and $\widehat\Sigma_\xi$ the moment estimates, from the training set, of $\operatorname{Cov}(a_i)$ and of the covariance of $\xi_{ij}$ pooled over participants (Appendix~\ref{app:optim}), the working estimate of $\psi_{ij}(\cdot)$ is
\begin{equation}
\widehat\psi_{ij}(t)=\widehat\psi_{ij}^{\mathrm{shr}}(t)+\rho_{ij}(t)\,\widehat\psi_{ij}^{\mathrm{avg}}(t),
\label{eq:pev}
\end{equation}
where
\[
\widehat\psi_{ij}^{\mathrm{shr}}(t)=\bm B^{(\mu)}(t)^{\top}(I-L_i)\widehat\Sigma_a(I-L_i)\bm B^{(\mu)}(t),\qquad
\widehat\psi_{ij}^{\mathrm{avg}}(t)=\frac{\bm B^{(\mu)}(t)^{\top}L_i\widehat\Sigma_\xi L_i\bm B^{(\mu)}(t)}{n_i-1}
\]
are computed once per fold. Because $\widehat\Sigma_\xi$ is pooled, the factor $\rho_{ij}(t)$ rescales the averaging noise to participant $i$: it is the ratio of the fitted residual variance at $t$ of participant $i$'s other trajectories to that of all trajectories, and it is the only part of~\eqref{eq:pev} that changes with the scale fit. Appendix~\ref{app:optim} defines $\rho_{ij}(\cdot)$, and Appendix~\ref{app:targets} states the approximation it involves.

\emph{Step 2: estimate the covariate slopes by conditional refitting.}
Subtracting $\widehat\psi_{ij}(t)$ from $Q_{ij}(t)$ could give negative responses, which the penalized Gamma fit with participant effects cannot use. We instead multiply $Q_{ij}(t)$ by an estimate of $\omega_{ij}(t)^2/\{\omega_{ij}(t)^2+\psi_{ij}(t)\}$, which by~\eqref{eq:pev-moment} would turn the mean of $Q_{ij}(t)$ into $\omega_{ij}(t)^2$. Because this factor involves the unknown $\omega_{ij}(t)^2$, we compute it from the current scale fit, refit the scale model to the rescaled $Q_{ij}(\cdot)$, and repeat; the number of refits $S$ is fixed in advance. Starting from the fitted predictor $\widehat\eta_{ij}^{(0)}(t)=X_{ij}^{\top}\widehat\alpha^{(0)}(t)+\widehat b_i^{(0)}(t)$ of~\eqref{eq:scale-working}, for $r=1,\ldots,S$, refit $r$ fits the model of Appendix~\ref{app:gamma}, with reselected penalties, to
\begin{equation}
\widetilde Q_{ij}^{(r)}(t)
=Q_{ij}(t)\,
\frac{\exp\{\widehat\eta_{ij}^{(r-1)}(t)\}}
{\exp\{\widehat\eta_{ij}^{(r-1)}(t)\}+\widehat\psi_{ij}^{(r-1)}(t)},
\label{eq:corrected-response}
\end{equation}
where $\exp\{\widehat\eta_{ij}^{(r-1)}(t)\}$ is the fitted residual variance from fit $r-1$ and $\widehat\psi_{ij}^{(r-1)}(t)$ is~\eqref{eq:pev} with $\rho_{ij}(t)$ computed from these fitted variances; refit $r$ gives $\widehat\eta_{ij}^{(r)}(t)=X_{ij}^{\top}\widehat\alpha^{(r)}(t)+\widehat b_i^{(r)}(t)$. In an idealized version of the update (Appendix~\ref{app:targets}), $\exp\{\widehat\eta_{ij}^{(r)}(t)\}$ decreases to $\omega_{ij}(t)^2$, and at each refit the error $\exp\{\widehat\eta_{ij}^{(r)}(t)\}-\omega_{ij}(t)^2$ is multiplied by at most $\psi_{ij}(t)/\{\omega_{ij}(t)^2+\psi_{ij}(t)\}<1$. With $\alpha(t)=\{\alpha_{\mathrm{int}}(t),\alpha_-(t)^{\top}\}^{\top}$ partitioned like $X_{ij}$, the corrected slopes are $\widehat\gamma_-(t)=\tfrac12\widehat\alpha_-^{(S)}(t)$.

\emph{Step 3: estimate the participant-averaged variance level.}
The covariate part $\exp\{x^{\top}\widehat\alpha^{(S)}(t)\}$ of the refits estimates the residual variance of a participant with $v_i(\cdot)=0$, and by~\eqref{eq:reference} $\bar q(x,t)$ is this variance times $\E\exp\{2v_i(t)\}$, which does not depend on $x$. By~\eqref{eq:pev-moment} and (C3), $\E\{Q_{ij}(t)-\psi_{ij}(t)\mid X_i\}=\bar q(X_{ij},t)+O(N^{-1})$. We therefore hold the slopes fixed and estimate from $Q_{ij}(t)-\widehat\psi_{ij}^{(S)}(t)$ a separate intercept function $\bar\alpha_{\mathrm{int}}(\cdot)$, the value of $\log\bar q(x,\cdot)$ at $x_-=0$. In this level fit, with $\bar\alpha_{\mathrm{int}}(\cdot)$ expanded in the scale basis and $\bar\eta_{ij}(t)=\bar\alpha_{\mathrm{int}}(t)+X_{ij,-}^{\top}\widehat\alpha_-^{(S)}(t)$, its $K_\omega$ coefficients solve
\begin{equation}
\sum_{i=1}^{N}\sum_{j=1}^{n_i}\sum_{k=1}^{T}
B_\ell^{(\omega)}(t_k)
\left[\frac{Q_{ij}(t_k)-\widehat\psi_{ij}^{(S)}(t_k)}
{\exp\{\bar\eta_{ij}(t_k)\}}-1\right]=0,
\qquad \ell=1,\ldots,K_\omega.
\label{eq:level}
\end{equation}
This is the Gamma quasi-score~\citep{wedderburn1974quasi} used as an estimating equation, so individual negative values of $Q_{ij}(t_k)-\widehat\psi_{ij}^{(S)}(t_k)$ are allowed. Appendix~\ref{app:optim} gives the starting values of the refits and the solver for~\eqref{eq:level}.

\emph{Corrected estimates.} With $\widehat{\bar\alpha}_{\mathrm{int}}(\cdot)$ solving~\eqref{eq:level}, the participant-averaged residual variance is estimated by
\[
\widehat{\bar q}(x,t)=\exp\{\widehat{\bar\alpha}_{\mathrm{int}}(t)+x_-^{\top}\widehat\alpha_-^{(S)}(t)\}.
\]
Participant-specific scale curves use the last refit, $\widehat v_i(t)=\tfrac12\widehat b_i^{(S)}(t)$ and $\widehat\omega_{ij}(t)=\exp[\tfrac12\{X_{ij}^{\top}\widehat\alpha^{(S)}(t)+\widehat b_i^{(S)}(t)\}]$; the level-fit intercept, which already averages over the $v_i(\cdot)$, is not substituted into them (Appendix~\ref{app:optim}).

\subsection{Participant-bootstrap inference}
\label{sec:bands}
Participants are independent, and every estimate in Sections~\ref{sec:estimation} and~\ref{sec:corrected} is built from participant-level contributions. Drawing $N$ participants with replacement therefore mimics drawing a new cohort from the same population, while keeping each participant's trajectories, their within-participant dependence and their number intact~\citep[Section~3.8]{davison1997bootstrap}. \citet{field2007cluster} and \citet{field2010robust} study such cluster bootstraps for clustered and multilevel data. Each replicate keeps each trajectory's original fold label and repeats all other lines of Algorithm1: the full-data mean fit, mean cross-fitting, Gamma penalty selection, the $S$ refits and the level fit. The bootstrap distribution therefore includes the variability these steps add.

For an estimand curve $\phi(\cdot)$, such as a coefficient function or $\log\bar q(x,\cdot)$, let $\widehat\phi^{*(b)}(\cdot)$ be its estimate from replicate $b=1,\ldots,B$ and $\mathrm{se}(t)$ the SD of $\widehat\phi^{*(1)}(t),\ldots,\widehat\phi^{*(B)}(t)$, the bootstrap standard error at $t$. The 95\% simultaneous band and the statistic that calibrates it are
\[
\widehat\phi(t)\pm c\,\mathrm{se}(t),\quad t\in\Tgrid,\qquad
Z_b^*=\max_{t\in\Tgrid}\frac{|\widehat\phi^{*(b)}(t)-\widehat\phi(t)|}{\mathrm{se}(t)},
\]
where $c$ is the empirical 0.95 quantile of $Z_1^*,\ldots,Z_B^*$~\citep{fan2000simultaneous,chang2017simultaneous,montiel2019simultaneous,cui2022fast}. Because $Z_b^*$ is the largest deviation of replicate $b$ from the estimate, in standard errors, 95\% of the replicate curves lie entirely inside the band. Each band is simultaneous over the grid of one curve only. A pointwise interval at $t$ uses instead the 0.95 quantile of the standardized deviations at $t$ alone. We report both: a pointwise interval concerns a single time, whereas a band supports statements about the whole curve, such as the times at which it excludes zero. For $\beta(\cdot)$, the bands cover its spline projection, the target that the score means identify under (C1) and (C3) (Appendix~\ref{app:targets}). For $\bar q(x,\cdot)$, work with $\log\bar q(x,\cdot)$ and exponentiate the limits. A whole-curve test of $\phi(t)=0$ for all $t\in\Tgrid$ compares the largest distance of the estimate from zero with $Z_1^*,\ldots,Z_B^*$:
\[
Z_{\mathrm{obs}}=\max_{t\in\Tgrid}\frac{|\widehat\phi(t)|}{\mathrm{se}(t)},\qquad
p=\frac{1+\#\{b:Z_b^*\geq Z_{\mathrm{obs}}\}}{B+1}\;\geq\;\frac{1}{B+1}.
\]

When testing multiple coefficient functions, such as the non-intercept terms in a model, we adjust the resulting $p$-values using the Benjamini--Hochberg (BH) procedure~\citep{benjamini1995fdr}. Appendix~\ref{app:optim} provides further details on the treatment of duplicated participants and failed fits in bootstrap replicates, the quantile definition, and the lower bound imposed on $\mathrm{se}(t)$.

%\subsection{Estimation algorithm}
%\label{sec:algorithm}
%Algorithm1 summarizes the estimation in Sections~\ref{sec:estimation} and~\ref{sec:corrected}.

\subsection{Software implementation}
\label{sec:software}
All analyses were performed in R version 4.6.1~\citep{R2026}. Cubic B-splines were generated with \texttt{splines::bs} and normalized by QR decomposition. The score-wise mean models and the penalized Gamma scale models were fitted with custom R functions that use the structure of the model: with the orthonormal basis, each curve model separates into score-wise problems, and each participant effect involves only that participant's trajectories. The mean fits minimize the profiled REML criterion of \texttt{lme4::lmer}~\citep{bates2015lme4}. The Gamma fits use the penalized quasi-likelihood iterations~\citep{breslow1993approximate} and fREML criterion of \texttt{mgcv::bam} with \texttt{method = "fREML"}~\citep{wood2015large,wood2017generalized}, evaluated exactly rather than with the covariate discretization of \texttt{discrete = TRUE}~\citep{wood2017gigadata}.

The prediction-error covariances, conditional refits, level fit and participant bootstrap were implemented in custom R functions. Bootstrap refits were run in parallel, with a separate random-number seed for each attempt, so that results do not depend on the number of workers. In the analysis of Section~\ref{sec:application}, Algorithm1 fits the penalized Gamma model~\eqref{eq:scale-working} four times: once initially and once in each of the three conditional refits of Step~2 in Section~\ref{sec:corrected} ($S=3$). On these data, one such fit took on average 0.36 seconds (SD 0.02) with our functions and 178.62 seconds (SD 1.30) with \texttt{mgcv::bam} (version 1.9-4), even with its faster approximate option \texttt{discrete = TRUE}. Both were timed over 10 single-thread runs on a MacBook Pro with an Apple M5 chip and 16 GB of memory. The 500 bootstrap refits therefore need 2{,}000 such fits, which at these speeds take about 12 minutes with our functions and about 99 hours with \texttt{mgcv::bam}.
\FloatBarrier
\FloatBarrier
\section{AEGIS case study}
\label{sec:aegis}
\label{sec:application}

\subsection{Data and analysis}

The A Estrada Glycation and Inflammation Study (AEGIS) investigated blood glucose and its relation to inflammation and obesity in the general adult population~\citep{matabuena2026beyond}. Its baseline examination took place in A Estrada, Galicia, northwestern Spain, between August 2012 and March 2015~\citep{calvomalvar2025age}. Adults aged 18 years or older were drawn as a stratified random sample from the Spanish National Health System registry~\citep{gude2017glycemic,matabuena2026beyond}. The study was approved by the Clinical Research Ethics Committee of Galicia (code 2012/025), and all participants gave informed consent. Of the 1,516 participants examined, 581 completed blinded CGM monitoring for up to seven days and, over the same days, recorded their meals in a food diary that a dietitian reviewed. Of these 581 participants, 516 did not have diabetes~\citep{matabuena2026beyond}. Among them, we analyzed the $N=363$ who had complete data and at least three recorded meals of each type (breakfast, lunch and dinner), 5,237 meals in total (9--15 per participant). The method itself needs fewer meals (Algorithm1); we required three of each type so that each participant's summaries in Section~\ref{sec:aegis-r2} rest on at least three meals of each type.

The glucose response of a person to the same meal can differ substantially between different days and conditions, which limits how well one response predicts the next~\citep{hengist2025imprecision}. Because each participant recorded repeated meals in free-living conditions, the expected postmeal course of glucose can be separated from this meal-to-meal variation. We therefore study how meal composition and participant characteristics are associated with the expected course and with this variation. Specifically, we address five questions. (i) Does the association of carbohydrate with the mean and the meal-to-meal variability of glucose differ between breakfast, lunch and dinner? (ii) How is premeal glucose associated with the mean and the variability? (iii) Is HbA1c associated with both the mean and the variability? (iv) How does the variability change after a meal and between meal types? (v) Is the individual $R^2$ similar across participants?

In the notation of Section~\ref{sec:model}, $j$ indexes the meals of participant $i$, and the fitted model~\eqref{eq:model} specifies
\[
\E\{Y_{ij}(t)\mid X_{ij},u_i,v_i\}=X_{ij}^{\top}\beta(t)+u_i(t),\qquad
\tfrac12\log\Var\{Y_{ij}(t)\mid X_{ij},u_i,v_i\}=X_{ij}^{\top}\gamma(t)+v_i(t),
\]
with the same covariates in both parts,
\[
X_{ij}=(1,\mathrm{age}_i,\mathrm{BMI}_i,\mathrm{HbA1c}_i,\mathrm{sex}_i,\mathrm{cal}_{ij},Y_{ij}(-5),I_{ij,\mathrm L},I_{ij,\mathrm D},\mathrm{carb}_{ij}I_{ij,\mathrm B},\mathrm{carb}_{ij}I_{ij,\mathrm L},\mathrm{carb}_{ij}I_{ij,\mathrm D})^{\top},
\]
where $Y_{ij}(-5)$ is premeal glucose, $\mathrm{cal}_{ij}$ and $\mathrm{carb}_{ij}$ are the calories (kcal) and carbohydrate (g) of meal $j$, and $\mathrm{sex}_i$ is 0 for female and 1 for male participants. Age, BMI, HbA1c and sex describe the participant, and the other covariates describe the meal. Table~\ref{tab:aegis-description} summarizes the covariates.

Carbohydrate intake is a primary dietary determinant of postprandial glucose, whereas total caloric intake reflects the overall energy content of a meal, including contributions from fat and protein. HbA1c reflects average glycemic exposure over the preceding months and is widely used in the diagnosis of diabetes and prediabetes~\citep{S14}. Premeal glucose represents the initial condition of the metabolic system from which the postprandial glucose response evolves, influencing the subsequent glucose trajectory~\citep{matabuena2026beyond}. Age, sex, BMI, and meal timing have also been associated with postprandial glycemic responses in this population~\citep{calvomalvar2025age}. Finally, meal-type indicators allow glucose responses and their associations with carbohydrate intake to differ across breakfast, lunch, and dinner.

The indicator $I_{ij,\mathrm B}$ is 1 if meal $j$ is a breakfast and 0 otherwise; $I_{ij,\mathrm L}$ and $I_{ij,\mathrm D}$ do the same for lunch and dinner. Breakfast is the reference meal, so only $I_{ij,\mathrm L}$ and $I_{ij,\mathrm D}$ appear on their own in $X_{ij}$, and their coefficients are differences from breakfast. The carbohydrate amount $\mathrm{carb}_{ij}$ enters only through the last three entries of $X_{ij}$, its products with the three indicators, so each meal type has its own carbohydrate slope. The difference between two meal types therefore depends on the carbohydrate amount, and we report it at 50 g with the other covariates fixed.

The fixed effects are the coefficient functions $\beta(\cdot)$ and $\gamma(\cdot)$, whose 11 non-intercept curves are shown in Figures~\ref{fig:aegis} and~\ref{fig:aegis-scale}. The random effects are the participant mean and scale effects $u_i(\cdot)$ and $v_i(\cdot)$, shared by all meals of participant $i$. Given $X_{ij}$, $u_i$ and $v_i$, the only random part of $Y_{ij}(t)$ is the standardized error $\epsilon_{ij}(t)$, which varies from meal to meal. The variance in the second equation therefore describes variation between meals of the same participant, and $\Var\{u_i(t)\}$ does not enter it. Variation between participants with the same covariates enters through $u_i(\cdot)$ and $v_i(\cdot)$. Their distribution is not specified, and by (C3) it does not depend on the covariates. Section~\ref{sec:aegis-reference} reports their estimated SDs. By (C1) and (C3), the reference mean $m(x,t)=\E\{Y_{ij}(t)\mid X_{ij}=x\}$ is the expected glucose $t$ minutes after a meal with covariates $x$, and the participant-averaged residual variance $\bar q(x,t)=\E[\{Y_{ij}(t)-\mu_{ij}(t)\}^2\mid X_{ij}=x]$ is the expected squared deviation of such a meal from its participant's mean curve.

Trajectories were analyzed on a grid $\Tgrid$ of $T=37$ points from 0 to 180 minutes after meal start, at five-minute intervals. We used $K_\mu=8$ mean and $K_\omega=6$ scale basis functions and applied the estimation procedure in Sections~\ref{sec:estimation}--\ref{sec:corrected}, with $F=15$ meal-wise cross-fitting folds formed by randomly ordering each participant's meals and $S=3$ conditional refits. Effects are reported in the natural increments shown in the figures. Carbohydrate slopes describe an increase of 10 g at fixed calories.

Pointwise intervals, simultaneous bands and whole-curve tests were computed as in Section~\ref{sec:bands} from $B=500$ participant-bootstrap refits, with BH adjustment within each model's 11 non-intercept coefficient curves. Appendix~\ref{app:application} gives implementation and contrast details.

\begin{table}[htbp]
\centering\small
\caption{Characteristics of the 363 participants and their 5,237 meals. Continuous variables are summarized as mean (SD); female participants are summarized as count (percentage). Participant characteristics are calculated once per participant, and meal covariates are calculated within each meal type.}
\label{tab:aegis-description}
\begin{tabular*}{0.65\textwidth}{@{\extracolsep{\fill}}lr@{}}
\toprule
\multicolumn{2}{l}{\textbf{(a) Participant characteristics}}\\
Characteristic & Summary\\
\midrule
Age (years) & 47.2 (14.2) \\
BMI (kg/m$^2$) & 28.1 (5.0) \\
HbA1c (\%) & 5.35 (0.33) \\
Female, n (\%) & 230 (63.4) \\
\bottomrule
\end{tabular*}

\vspace{0.6em}
\begin{tabular*}{\textwidth}{@{\extracolsep{\fill}}lrrrr@{}}
\toprule
\multicolumn{5}{l}{\textbf{(b) Meals}}\\
Meal type & Meals & Calories (kcal) & Carbohydrate (g) & $Y(-5)$ (mg/dL)\\
\midrule
Breakfast & 1727 & 318.1 (200.1) & 44.8 (29.1) & 102.6 (13.9) \\
Lunch & 1755 & 879.6 (425.6) & 74.0 (46.0) & 100.8 (14.7) \\
Dinner & 1755 & 649.2 (389.9) & 59.2 (40.3) & 104.6 (16.2) \\
\bottomrule
\end{tabular*}
\end{table}

\subsection{Covariate associations in conditional mean and conditional variability}
\label{sec:aegis-associations}
\begin{figure}[htbp]
\centering
\includegraphics[width=\textwidth]{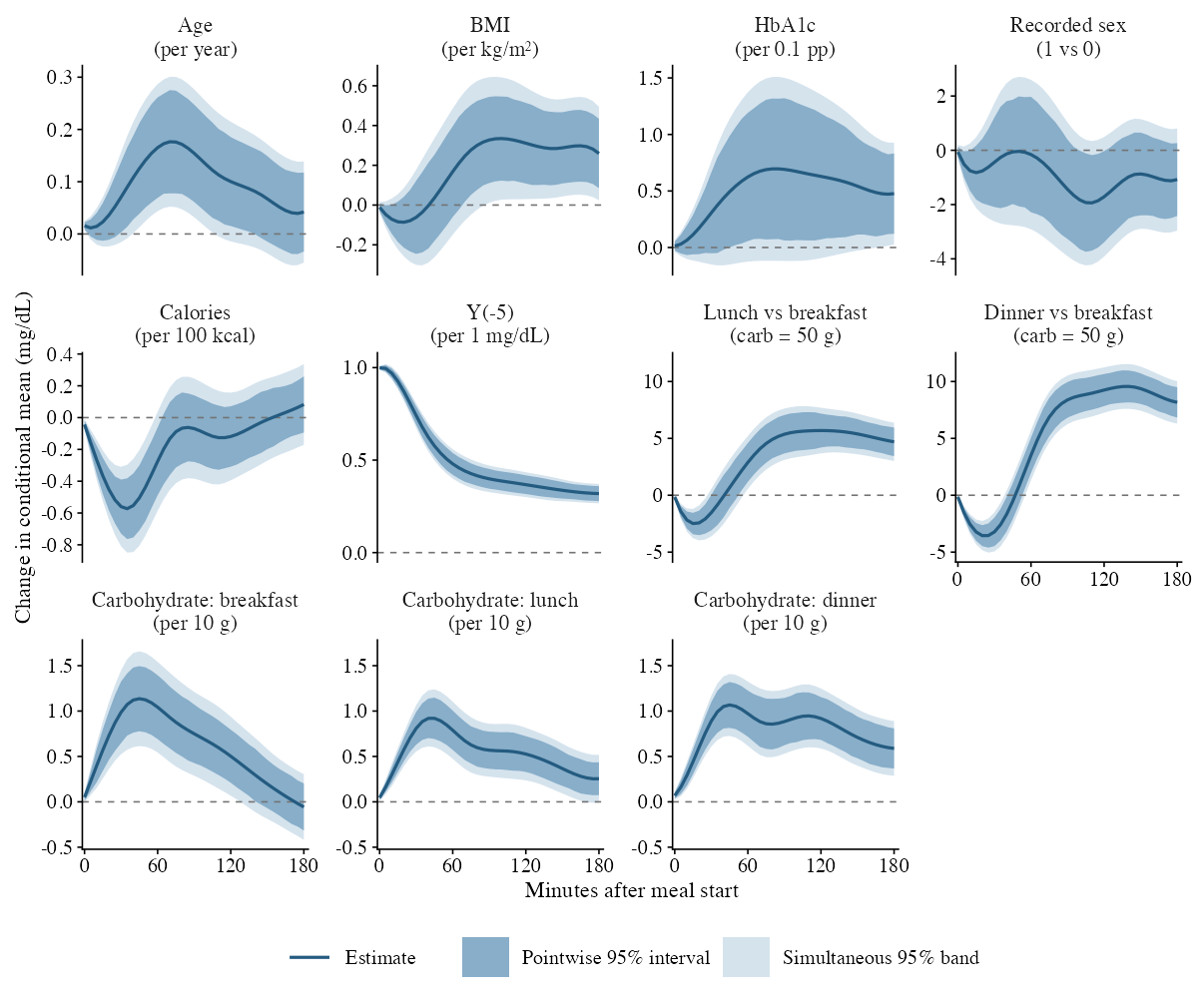}
\caption{Mean-glucose associations over postmeal time and meal type. Panels show 11 conditional-mean coefficient effects in mg/dL, with covariate increments in the labels. Dark shading shows pointwise nominal 95\% intervals and light shading simultaneous bands from 500 participant-bootstrap refits. Meal contrasts compare lunch and dinner with breakfast at 50 g carbohydrate; female (sex coded 0) is the recorded-sex reference. Calories and carbohydrate adjust for each other. The three carbohydrate panels share a vertical scale, as do the two meal-contrast panels.}
\label{fig:aegis}
\end{figure}

\begin{figure}[htbp]
\centering
\includegraphics[width=\textwidth]{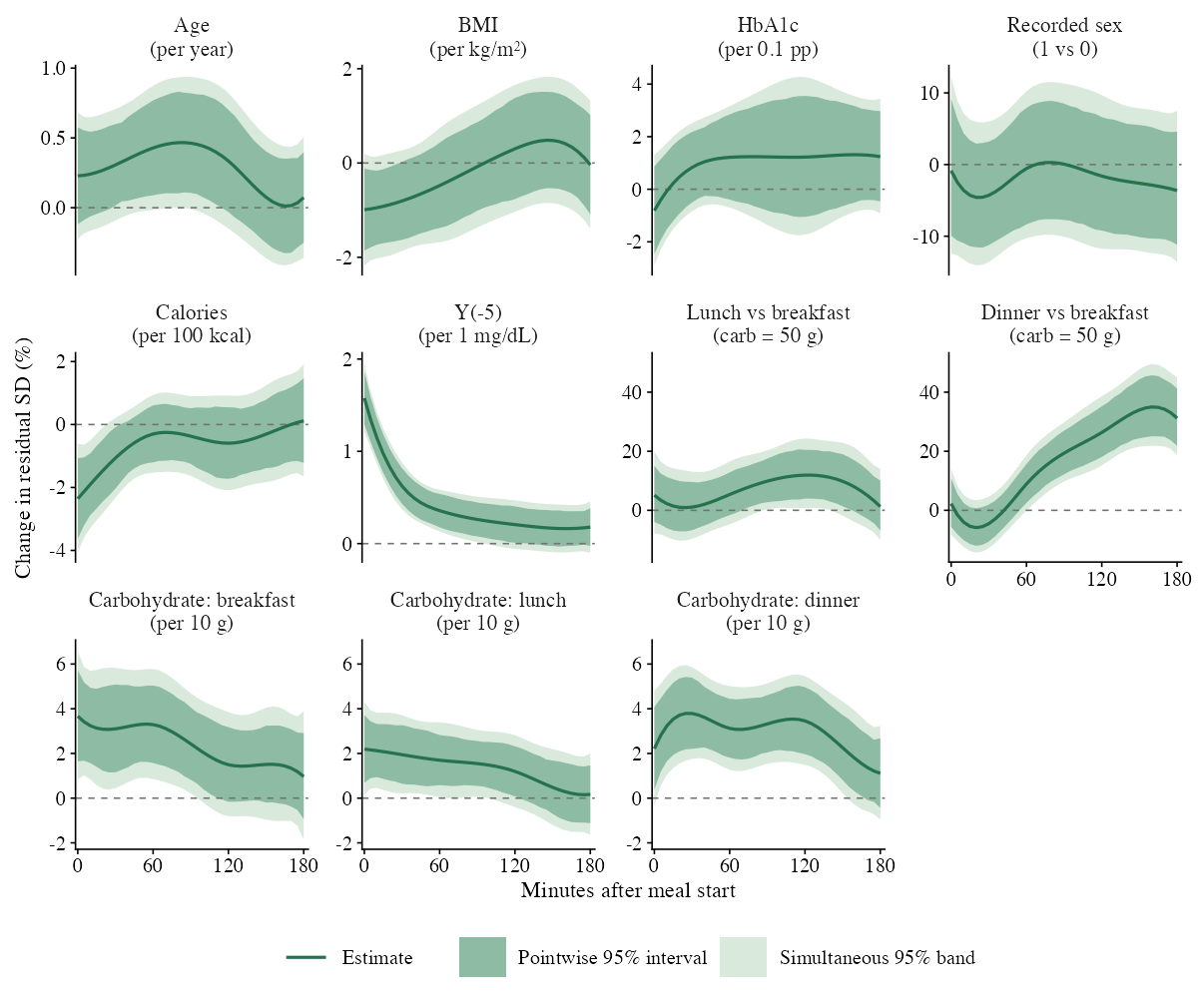}
\caption{Residual-SD associations over postmeal time. Effects are displayed as $100\{\exp(d_a\gamma_a(t))-1\}$ percent change in residual SD, with increments $d_a$ in the panel labels; effects and band limits are calculated on the log-SD scale before transformation. Dark and light shading show pointwise nominal 95\% intervals and simultaneous bands from 500 participant-bootstrap refits. Meal contrasts use breakfast at 50 g carbohydrate, and female (sex coded 0) is the recorded-sex reference. The three carbohydrate panels share a vertical scale, as do the two meal contrasts; each simultaneous band covers one coefficient curve.}
\label{fig:aegis-scale}
\end{figure}

Ten of the 11 mean coefficient curves and seven of the 11 log-SD curves had BH-adjusted $p<0.05$ (Figures~\ref{fig:aegis} and~\ref{fig:aegis-scale}, Table~\ref{tab:aegis}).

Question~(i) concerns carbohydrate and meal type. Carbohydrate was associated with higher mean glucose and greater residual SD at all three meals. The slopes differed between meals only in contrasts involving dinner, and only after 100 minutes. In the mean model, the carbohydrate bands excluded zero at 5--125 minutes for breakfast, 0--165 for lunch and 0--180 for dinner. In the log-SD model, they excluded zero at 0--95 minutes for breakfast, 0--60 and 75--95 for lunch, and 5--150 for dinner.

We also compared the three carbohydrate slopes directly, using paired contrasts within each bootstrap refit. The dinner-minus-breakfast mean slope was positive at 135--180 minutes and passed BH adjustment over the three meal-pair comparisons ($p_{\mathrm{BH}}=0.012$). The dinner-minus-lunch slope was positive at 110--135 and 155--170 minutes in the mean model ($p_{\mathrm{BH}}=0.045$) and at 100--140 minutes in the log-SD model ($p_{\mathrm{BH}}=0.048$). The other comparisons did not pass the adjusted 0.05 threshold (Appendix Table~\ref{tab:aegis-carb-contrasts}).

Higher premeal glucose (question~ii) was associated with higher mean glucose over the whole 180 minutes and with greater residual SD at 0--95 minutes. Premeal glucose is likely related to the participant effects, contrary to (C3), so these associations may partly reflect differences between participants.

For HbA1c (question~iii), the evidence was weak. HbA1c was associated with the mean only at 170--180 minutes ($p_{\mathrm{BH}}=0.033$). Its association with the residual SD did not pass BH adjustment ($p_{\mathrm{BH}}=0.329$). Participants did not have diabetes, so HbA1c varied little (SD 0.33 percentage points; Table~\ref{tab:aegis-description}).

Among the other covariates, age had a positive mean association at 40--105 minutes, with an isolated finding at 0 minutes. BMI had a positive mean association at 90--180 minutes. With carbohydrate held fixed, calories had a negative association with the mean at 0--55 minutes and with residual SD at 0--20 minutes. Recorded sex passed BH adjustment in neither model, nor did age and BMI in the log-SD model. Section~\ref{sec:aegis-reference} describes the meal-type contrasts.

The directions of the mean associations agree with those reported by \citet{calvomalvar2025age}, who applied multilevel functional regression to 514 participants without diabetes from the same study but only focus on conditional mean estimation. In both analyses, glucose increased with meal carbohydrate (glycemic load in their analysis), age, BMI and HbA1c, and the carbohydrate association lasted longer after lunch and dinner than after breakfast. 

\begin{table}[htbp]
\centering\small
\setlength{\tabcolsep}{3pt}
\caption{AEGIS whole-curve tests and exclusion windows. BH adjustment is applied within each model's 11 coefficient curves. Values of $p_{\mathrm{BH}}$ below 0.05 are in bold. Windows are in minutes and use each curve's nominal 95\% simultaneous band, with signs in parentheses; they are not jointly simultaneous across coefficients. Covariate increments follow Figures~\ref{fig:aegis}--\ref{fig:aegis-scale}, and meal contrasts are evaluated at 50 g carbohydrate.}
\label{tab:aegis}
\begin{tabularx}{\textwidth}{@{}>{\raggedright\arraybackslash}p{0.24\textwidth}c>{\raggedright\arraybackslash}Xc>{\raggedright\arraybackslash}X@{}}
\toprule
 & \multicolumn{2}{c}{Conditional mean, $\beta_a(t)$} & \multicolumn{2}{c}{Log residual SD, $\gamma_a(t)$}\\
\cmidrule(lr){2-3}\cmidrule(l){4-5}
Coefficient & $p_{\mathrm{BH}}$ & Windows & $p_{\mathrm{BH}}$ & Windows\\
\midrule
Age & \textbf{0.003} & 0 (+); 40--105 (+) & 0.074 & none \\
BMI & \textbf{0.017} & 90--180 (+) & 0.117 & none \\
HbA1c & \textbf{0.033} & 170--180 (+) & 0.329 & none \\
Recorded sex & 0.124 & none & 0.707 & none \\
Calories & \textbf{0.003} & 0--55 ($-$) & \textbf{0.004} & 0--20 ($-$) \\
Premeal glucose & \textbf{0.003} & 0--180 (+) & \textbf{0.004} & 0--95 (+) \\
Lunch vs breakfast & \textbf{0.003} & 5--25 ($-$); 60--180 (+) & \textbf{0.038} & 90--140 (+) \\
Dinner vs breakfast & \textbf{0.003} & 5--35 ($-$); 60--180 (+) & \textbf{0.004} & 65--180 (+) \\
Carbohydrate: breakfast & \textbf{0.003} & 5--125 (+) & \textbf{0.004} & 0--95 (+) \\
Carbohydrate: lunch & \textbf{0.003} & 0--165 (+) & \textbf{0.015} & 0--60 (+); 75--95 (+) \\
Carbohydrate: dinner & \textbf{0.003} & 0--180 (+) & \textbf{0.004} & 5--150 (+) \\
\bottomrule
\end{tabularx}
\end{table}
\FloatBarrier

\subsection{Reference mean and variability}
\label{sec:aegis-reference}
\begin{figure}[htbp]
\centering
\includegraphics[width=\textwidth]{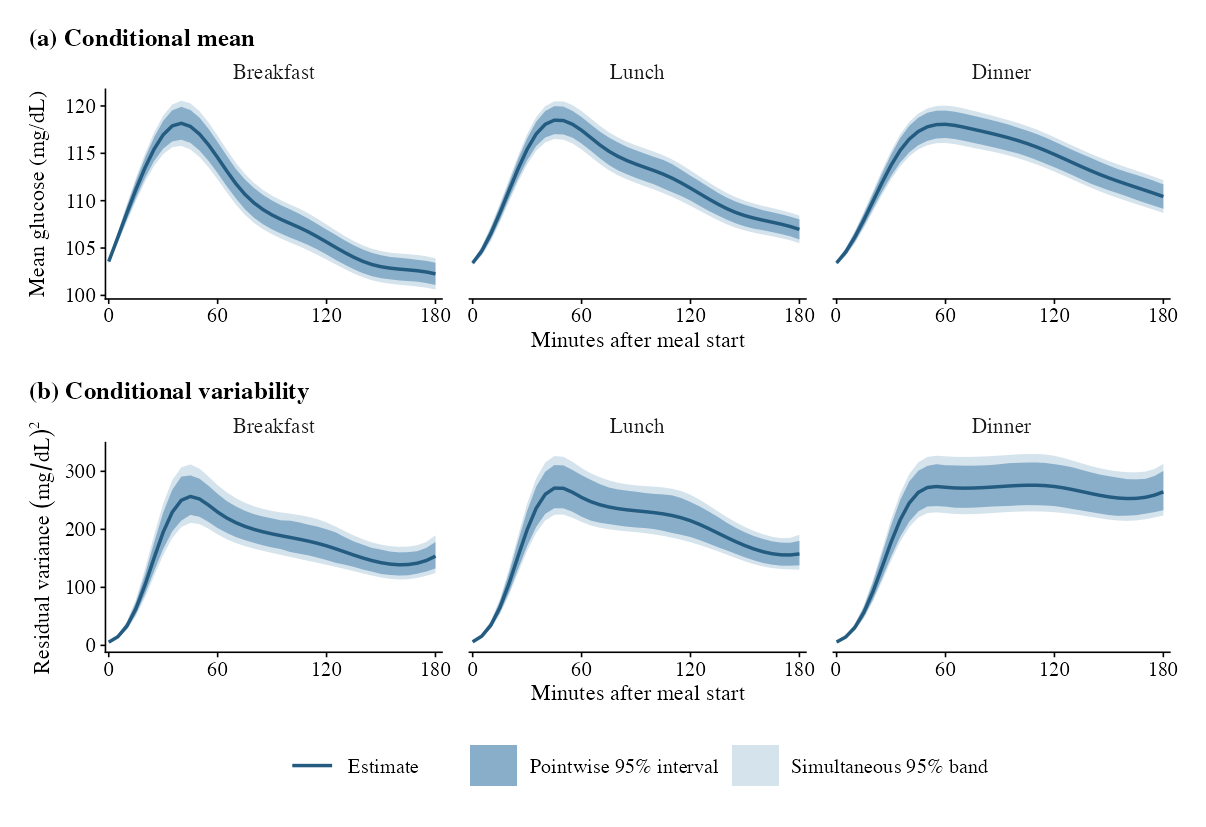}
\caption{Dinner has a broader late elevation in the reference mean and residual variance. Part (a) shows $m(x,t)$ and part (b) shows $\bar q(x,t)$ at the sample covariate means, female sex and 50 g carbohydrate. The participant-averaged residual variance excludes between-participant mean variation. The three meal panels within each part share a vertical scale. Dark and light shading show pointwise nominal 95\% intervals and simultaneous bands from 500 participant-bootstrap refits; variance limits are constructed on the log scale and exponentiated.}
\label{fig:aegis-reference}
\end{figure}

To compare the fitted trajectories on a common covariate profile, we fixed the continuous covariates at their sample means, sex at female and carbohydrate at 50 g. Figure~\ref{fig:aegis-reference} shows, at this profile $x$, the reference mean $m(x,t)$ and the participant-averaged residual variance $\bar q(x,t)$ defined in Section~\ref{sec:targets}. The mean peaks occurred at 40, 45 and 60 minutes for breakfast, lunch and dinner, respectively, with peak values of 118.1--118.5 mg/dL. At this profile, the differences between the lunch or dinner curves and the breakfast curve are the meal contrasts tested in Table~\ref{tab:aegis}: the mean contrast for $m(x,\cdot)$, and twice the log-SD contrast for $\log\bar q(x,\cdot)$. In the mean, lunch and dinner were below breakfast at 5--25 and 5--35 minutes, respectively, and above breakfast at 60--180 minutes. Dinner fell most slowly after its peak.

Turning to question~(iv), $\bar q(x,t)$ rose from about 6 to 257--271 (mg/dL)$^2$ by 45 minutes, then fell after breakfast and lunch but not after dinner. The value at meal start is low because premeal glucose is a covariate. By 180 minutes, $\bar q(x,t)$ was about 40\% below its peak after breakfast and lunch, while the dinner curve stayed between 253 and 276 (mg/dL)$^2$ from 45 minutes on. In the log-SD meal contrasts, lunch had greater residual SD than breakfast at 90--140 minutes, and dinner at 65--180 minutes. No lunch-versus-dinner contrast was computed at this profile.

Participants differed substantially in residual variability: averaged over 5 to 180 minutes, the model-implied SD of $v_i(t)$ was 0.35, so at the same covariates, participants one SD apart differed in residual SD by a factor of about 1.4. Averaged over the grid, the participant mean effect $u_i(t)$ had SD 6.9 mg/dL and accounted for 12--22\% of the variance of the projection scores $C_{ij\ell}$, so most of the conditional variance was meal-to-meal variation within a participant.

\FloatBarrier

\subsection{Individual contribution of participant mean effects}
\label{sec:aegis-r2}

An individual $R^2$ summarizes how much model-based squared error remains for participant $i$ relative to a common population variance. A meal $j$ of participant $i$ can be predicted by the reference mean $m(X_{ij},t)=X_{ij}^{\top}\beta(t)$, which uses only the covariates, or by the participant-specific mean $\mu_{ij}(t)=m(X_{ij},t)+u_i(t)$. Averaging their expected squared errors over 0--180 minutes gives
\begin{equation}
\begin{aligned}
R_{i,\mathrm{fixed}}^2(X_{ij})&=1-\frac{1}{D}\,\frac{1}{180}\,\E\|Y_{ij}(\cdot)-m(X_{ij},\cdot)\|_{\mathcal H}^2,\\
R_{i,\mathrm{participant}}^2(X_{ij})&=1-\frac{1}{D}\,\frac{1}{180}\,\E\|Y_{ij}(\cdot)-m(X_{ij},\cdot)-u_i(\cdot)\|_{\mathcal H}^2,\\
D&=\frac{1}{180}\int_{\mathbb T}\Var\{Y_{ij}(t)\}\,dt.
\end{aligned}
\label{eq:aegis-r2}
\end{equation}
In the first two lines, the expectation is over the standardized error $\epsilon_{ij}(\cdot)$, which varies from meal to meal, with $X_{ij}$, $u_i(\cdot)$ and $v_i(\cdot)$ fixed, so by (C1) the expected squared errors at time $t$ are $\omega_{ij}(t)^2+u_i(t)^2$ and $\omega_{ij}(t)^2$. The variance in the third line is over participants and meals. We estimate $D$ by the average of $\frac{1}{180}\|Y_{ij}(\cdot)-\bar Y(\cdot)\|_{\mathcal H}^2$ over all $M=5{,}237$ curves, where $\bar Y(t)$ is their pointwise mean, which gives $D=361.5$ (mg/dL)$^2$, an SD of 19.0 mg/dL; integrals use normalized trapezoidal weights on $\Tgrid$. We report the averages over the $n_i$ meals of participant $i$, written $R_{i,\mathrm{fixed}}^2$ and $R_{i,\mathrm{participant}}^2$. Both use all covariates and $v_i(\cdot)$ and differ only by the paired contribution of $u_i(\cdot)$, $\Delta R_i^2=R_{i,\mathrm{participant}}^2-R_{i,\mathrm{fixed}}^2\ge0$. They are estimated by substituting $\widehat u_i(\cdot)$ and $\widehat\omega_{ij}(\cdot)$, and are structural model-based summaries rather than measures of observed prediction accuracy.

\begin{figure}[ht!]
\centering
\includegraphics[width=\textwidth]{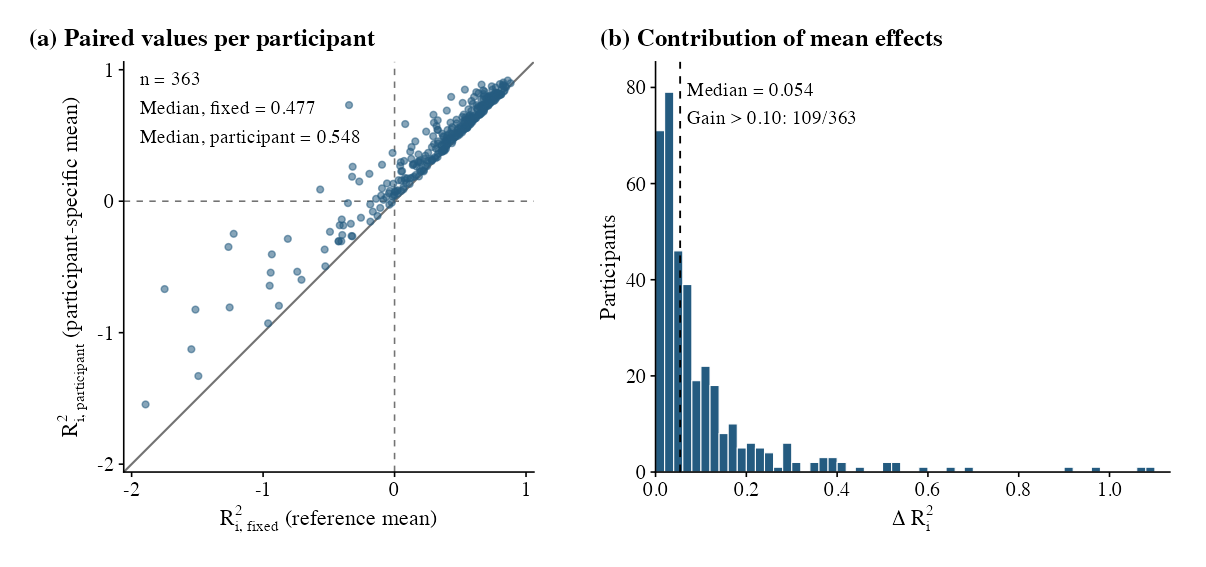}
\caption{Participant mean effects reduce model-based squared error by a median of 5.4\% of the common population variance ($D=361.5$ (mg/dL)$^2$). (a) Each point pairs $R_{i,\mathrm{fixed}}^2$ and $R_{i,\mathrm{participant}}^2$ of one participant; axes have the same range and the solid line is the identity line. (b) Distribution of the paired contribution $\Delta R_i^2$, with its median marked. All 363 participants and negative values are retained. Both values retain the fitted participant scale effect and treat participant effects as known; the nonnegative contribution follows from the definition. No individual uncertainty intervals are shown.}
\label{fig:individual-r2}
\end{figure}

The individual $R^2$ varied widely between participants, so the answer to question~(v) is no. The median $R_i^2$ was 0.477 (IQR 0.173--0.639) for the reference mean and 0.548 (0.294--0.702) for the participant-specific mean (Figure~\ref{fig:individual-r2}a). The median paired contribution was 0.054 (IQR 0.024--0.117), and 109 of 363 participants had $\Delta R_i^2>0.1$ (panel b). This median, 5.4\% of $D$, is not the difference between the two group medians. There were 57 negative values of $R_{i,\mathrm{fixed}}^2$ and 38 of $R_{i,\mathrm{participant}}^2$, where the model-based squared error exceeded the common baseline.

\FloatBarrier
\section{Discussion}
\label{sec:discussion}
\label{sec:pending}

Previous work has used multilevel functional regression to study the conditional mean of postprandial glucose trajectories in the AEGIS cohort, leaving the analysis of conditional variability for future research~\citep{matabuena2026beyond}. In this paper, we extend this approach by jointly modeling the conditional mean and variance of repeated functional observations. To our knowledge, this is the first multilevel functional location--scale model that allows covariates and participant-specific random effects to influence both the mean and residual variability of the trajectories.

The AEGIS results show the value of studying both the mean and variability of postprandial glucose trajectories. At the reference covariate profile, the participant-averaged residual variance reached 257--276 (mg/dL)$^2$, corresponding to a standard deviation of about 16--17 mg/dL. This is similar in magnitude to the average postmeal glucose increase of approximately 15 mg/dL. Higher carbohydrate intake was associated with greater glucose variability at breakfast, lunch, and dinner. We also found that residual variability was higher after dinner than after breakfast, particularly from 65 minutes after the meal onwards. Variability also differed considerably between participants. For the same covariate values, a one-standard-deviation difference in the participant-specific scale effect corresponded to an approximately 1.4-fold difference in residual standard deviation. These findings represent associations rather than causal effects. Finally, the contribution of participant-specific mean effects to individual $R^2$ varied across participants. These $R^2$ values describe how well the model explains individual trajectories relative to a common population variance and should not be interpreted as measures of predictive performance.

Several limitations should be acknowledged. Our interpretation relies on condition (C3), which may be restrictive because premeal glucose is likely associated with participant-specific effects in AEGIS. Although the model allows the mean and scale random effects to be correlated, as in conventional mixed-effects location--scale models~\citep{hedeker2008application}, the proposed prediction-error correction does not explicitly account for this correlation. Moreover, the residual variability observed in AEGIS may reflect several sources beyond physiological variation, including sensor measurement error, unrecorded meal components, and inaccuracies in carbohydrate intake or meal timing. Some of these sources may increase with meal size and contribute to the observed associations between carbohydrate intake and residual variability. The proposed correction also relies on an approximation of the prediction-error variance through the rescaling factor $\rho_{ij}(\cdot)$ and does not account for participant-level variation outside the selected mean basis. The confidence bands do not correct for smoothing or projection bias~\citep{hall2013simple}.

Future work could extend the proposed correction to account for correlation between participant-specific mean and scale effects and explore other aspects of the conditional distribution beyond the mean and variance. More generally, the proposed framework provides a way to study both the expected response and its variability in repeated functional data, with potential applications to other wearable sensors and longitudinal physiological measurements.

\label{body-end}

\begingroup
\small
\setlength{\bibsep}{3pt}
\setlength{\bibhang}{0.8em}
\bibliographystyle{plainnat}
\bibliography{references}
\endgroup
\end{document}